# Self-Supervised Contextual Language Representation of Radiology Reports to Improve the Identification of Communication Urgency

**Xing Meng, MS[1], Craig H. Ganoe, MS[2], Ryan T. Sieberg, MD[3],**
**Yvonne Y. Cheung, MD[3], Saeed Hassanpour, PhD[1,2,4]**

**[1]Computer Science Department, Dartmouth College, Hanover, NH 03755, USA;**
**[2]Biomedical Data Science Department, Dartmouth College, Hanover, NH 03755, USA;**
**[3]Radiology Department, Dartmouth-Hitchcock Medical Center, Lebanon, NH 03756, USA;**
**[4]Epidemiology Department, Dartmouth College, Hanover, NH 03755, USA**

## Abstract

*Machine learning methods have recently achieved high-performance in biomedical text analysis. However, a major bottleneck in the widespread application of these methods is obtaining the required large amounts of annotated training data, which is resource intensive and time consuming. Recent progress in self-supervised learning has shown promise in leveraging large text corpora without explicit annotations. In this work, we built a self-supervised contextual language representation model using BERT, a deep bidirectional transformer architecture, to identify radiology reports requiring prompt communication to the referring physicians. We pre-trained the BERT model on a large unlabeled corpus of radiology reports and used the resulting contextual representations in a final text classifier for communication urgency. Our model achieved a precision of 97.0%, recall of 93.3%, and F-measure of 95.1% on an independent test set in identifying radiology reports for prompt communication, and significantly outperformed the previous state-of-the-art model based on word2vec representations.*

## Introduction

There has been a large amount of effort to apply machine learning methods to the domain of biomedical text analysis and information extraction[1–3]. Before the use of deep learning models, machine learning based Natural Language Processing (NLP) pipelines required feature engineering and feature extraction, which could be the most time-consuming parts of building machine learning models[4–7]. These feature engineering techniques were able to make machine learning algorithms more effective by transforming the training data into indicative features, and augmenting the data with additional information[8].

More recent deep learning models can obtain high performance across a wide variety of NLP tasks[9–11] without manual feature extraction. Although deep learning has eliminated the need for feature engineering, deep models require large amounts of labeled training data to be effective. Acquiring the required amounts of labeled data and annotations for developing deep learning models is particularly challenging in clinical domains where there are not many high-quality open-access labeled data available (due to concerns about patients' privacy) and domain expertise is scarce and expensive.

As one solution, self-supervised learning paradigms have been used in developing deep learning models to remedy the growing need for using deep learning models with large annotated training datasets. In contrast to unsupervised learning, self-supervised approaches still rely on input labels to optimize the training objective. However, unlike supervised and semi-supervised methods, these labels are automatically generated by exploiting the relations between different parts of input data, without human or expert involvement[12]. As an example of these self-supervised paradigms in NLP, deep learning models rely on dense vector representation for each word in input text (i.e., word embeddings), which can be automatically learned from the word co-occurrence patterns in a large corpus. The word2vec[13] model is one of the most widely-used methods for generating word embeddings from free texts. The word embeddings learned by word2vec models have been shown to carry semantic meanings and are useful in various NLP tasks[14]. Many biomedical deep-learning NLP systems have been built based on the word2vec word embeddings to represent input text[15–17].

Despite their many applications, most common word representation models, such as word2vec, are context-free and generate a single word embedding representation for each word in the vocabulary. Thus, the same word has the same representation, despite its various possible semantics in different contexts. As a result, models such as word2vec do not exploit the full power of the context for text semantic representation and self-supervised learning.

On the other hand, contextual language representation models generate the word embeddings based on the word context in the input text. BERT (Bidirectional Encoder Representations from Transformers)[18] is one of these contextual language representation models that has shown a lot of promise in NLP. For instance, BioBERT[19] has leveraged BERT for the three tasks of named entity recognition, relation extraction, and question answering in the biomedical domain. As another example, BertNet[20] has combined BERT language representations and convolutional neural networks (CNNs) and highly improved the performance of a question-answering model. Of note, there have been other independent efforts, in parallel to our work, to apply BERT models to clinical applications[21-22].

Delays in communicating urgent clinical findings in radiology exams to referring physicians account for major adverse patient outcomes every year[23]. The lack of clear guidelines on the radiological findings that require prompt communication between radiologists and referring physicians is one of the main reasons for such delays[24]. In this paper, we pre-trained and fine-tuned the BERT architecture on a large radiology report corpus, aiming to improve the identification of the radiology reports that require a radiologist's prompt communication to the referring physician in our existing NLP pipeline[25]. We integrated the pre-trained BERT model with our task-specific text classifier for prompt communications, and we evaluated and compared the performance of this new model with the previous state-of-the-art model using word2vec representations that were pre-trained on the same large radiology report corpus.

## Materials and Methods

In this section, we describe our dataset and the details of the proposed NLP pipeline that utilizes self-supervised contextual language representations to identify communication urgency in radiology. In addition, we outline our experiment to compare the proposed NLP to the previous state-of-the-art model for this task. An overview of our methodology in this study is shown in Figure 1.

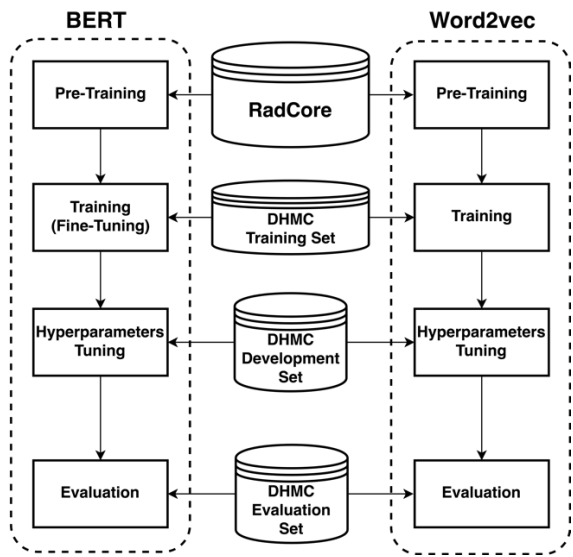

**Figure 1.** Methodology overview.

### Datasets

We utilized two sources of free-text radiology reports in the construction of our contextual language model: RadCore and DHMC. The RadCore dataset[6], the larger of our sources, contains about 2 million radiology reports that were deidentified and aggregated in 2007 from three major healthcare organizations: Mayo Clinic, MD Anderson Cancer Center, and Medical College of Wisconsin. The DHMC dataset contains 2,124 radiology reports that were extracted from our own institution, Dartmouth-Hitchcock Medical Center (DHMC), a tertiary academic care center in Lebanon,

New Hampshire. To balance the number of reports requiring prompt communication with those that do not, half the DHMC radiology reports extracted had "unexpected" or "communicated" tags in the picture archiving and communication system (PACS) at our institution, while the other half were selected randomly from those remaining. After extracting the reports, these tags from PACS were not used further in language modeling or developing the classifier. This study and the use of human subject data in this project were approved by the Dartmouth-Hitchcock Health institutional review board (D-HH IRB) with a waiver of informed consent.

Two radiologists (YYC and RTS) manually annotated the DHMC database radiology reports to establish ground truth labels for these reports. During this manual annotation, a binary label was assigned to each individual report indicating whether a prompt communication was needed on the basis of critical or unexpected radiological findings in that report. The agreement rate between our radiologist annotators in this task was 99.5%. Disagreements between the annotators on the need for prompt communication in a report were resolved through further discussions between annotators in an adjudication process. The resulting, labeled DHMC dataset was balanced across positive and negative radiology reports for prompt communication after our overlapping annotation, as we expected with our extraction approach, with 1,049 cases labeled as positive for prompt communication and 1,075 cases labeled as negative for prompt communication.

*Pre-training Dataset:* For our initial large corpora, we utilized the freely-avilable pre-trained BERT[18] model (pre-trained on Wikipedia and Google books corpus) and the word2vec model[13] (pre-trained on Google news dataset) for our baseline. We continued pre-training these publicly available models on the RadCore dataset. We extracted all radiology reports from the RadCore database which contained discernible "impression" section headings in the reports, resulting in 31,621 reports for pre-training of our self-supervised BERT model. We identified the impression section as the most relevant part of a radiology report for the computationally expensive language modeling in BERT pre-training. To pre-train our window-based word2vec model, we used all of the text in each report of the entire ~2 million report RadCore database, as this process was less computationally expensive.

*Training, Development, and Evaluation Datasets:* We randomly split the DHMC database into three parts, 60% (629 positive, 645 negative) for a training/fine-tuning dataset, 20% (210 positive, 215 negative) for a development set, and 20% (210 positive, 215 negative) for a held-out evaluation dataset. The training dataset (known as the fine-tuning dataset in BERT terminology) was used to fit the parameters of the model (e.g., weights of neurons in neural networks). The development set was used for tuning the model's hyperparameters (e.g., learning rate and number of training epochs). Finally, the held-out evaluation set was used for an unbiased evaluation for the model. Table 1 summarizes the datasets that are used in our study.

**Table 1.** The description of the radiology report datasets that were used in pre-training, training (fine-tuning), development, and evaluation of our models in this study.

| Dataset | Number of Radiology Reports | Source | Number of Labeled Radiology Reports | Utlized Radiology Report Section |
|---|---|---|---|---|
| Pre-Training for BERT | 31,621 | RadCore | Not labeled | Impressions section |
| Pre-Training for word2vec | ~2 million | | | All text |
| Training(Fine-Tuning) | 1,274 | DHMC | 629 positive 645 negative | Impressions section |
| Development | 425 | | 210 positive 215 negative | |
| Evaluation | 425 | | 210 positive 215 negative | |

### Contextual Language Model

BERT (Bidirectional Encoder Representations from Transformers) is a language representation model introduced by Google AI[26] in 2018. BERT is trained on a large corpus of Wikipedia articles and Google books to learn the masked language model. In this task, a randomly selected subset of words from the input is masked from the model. During training, BERT learns to guess masked words in input text, based on the context of the missing words and their relationships to context words. BERT architecture utilizes deep bidirectional transformers[27], a popular self-attention model that learns contextual relations between words in a text. The utilized transformers and the self-attention mechanism represent each word as a weighted sum of the word and its context words. This is in contrast to the traditional recurrent neural network (RNN) language models, which analyze a text sequence sequentially from left to right or concurrently from both directions.

Masked language modeling in BERT utilizes a self-supervised training framework and is trained on large unlabeled text corpora. The pre-trained BERT model generates word embeddings which then can be used for word representations in other NLP tasks. In training other NLP models that utilize BERT embeddings, the word representations are first initialized using the pre-trained BERT model. Subsequently, all the model parameters are fine-tuned on a labeled training set for the supervised downstream task. The utilization of BERT contextual word embeddings in common NLP tasks, such as question answering and sentiment analysis, has resulted in substantial performance improvements compared to training those models on other word embeddings[18].

In this work, we first pre-trained an existing BERT model from Google AI for contextual language modeling on our unlabeled RadCore pre-training dataset. The pre-trained BERT model was then used as the basis of a classification model on the labeled DHMC training dataset. During training, the last hidden layer of the pre-trained BERT model was fine-tuned for our classification task. We used our labeled development dataset to further optimize the hyperparameters.

*Pre-training the BERT Model:* Instead of pre-training the BERT model from scratch on the RadCore dataset, we utilized the pre-trained weights provided by Google AI group in October 2018. Pre-training the BERT masked language model was performed on the text of the impression sections from the radiology reports in the unlabeled RadCore pre-training dataset.

*Fine-tuning the BERT Model:* Fine-tuning of the BERT model was performed using the labeled DHMC fine-tuning dataset. Our classification task for fine-tuning BERT was set up similar to the binary single-sentence classification in the Corpus of Linguistic Acceptability (CoLA) task as described by Devlin et al.[18]. The goal of our classification task was to predict whether a report requires prompt communication to referring physicians or not, based on the given impression section of a radiology report. In this classification, we treat the impression section of each radiology report as a single sequence of text, and input it to the BERT model.

*Hyperparameters Tuning*: We explored a range of hyperparameters to optimize the BERT model for our classification task based on our development dataset. The distribution of impression section lengths in the radiology reports for the RadCore and DHMC datasets is shown in Figure 2. Through our hyperparameter tuning, we observed the optimal max sequence length is 128 for our task.

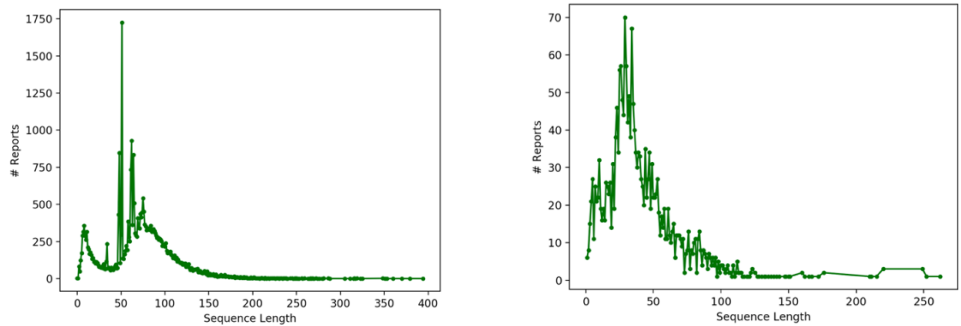

**Figure 2.** Distribution of impression section length (in words) for the RadCore dataset (left) and the DHMC dataset (right).

We also explored the optimal batch size, learning rate, and number of training epochs in our hyperparameter tuning process. In this process, we performed an exhaustive grid search over these hyperparameters and chose the parameters that performed best on the development set. Particularly, to find an optimal learning rate we performed a two level grid search. In the first level, we ran a grid search with a 0.1 step size to find the optimal learning rate of 1e-5 using a low resolution search (shown in Figure 3 left). In the next level, we performed a finer grained grid search around the optimal value from the previous level by using a smaller step size (shown in Figure 3 right). After this two level search, we observed that 2e-5 was the optimal value for learning rate in our task. Figure 4 shows the optimal number of epochs for training our model is 5, as our model starts to overfit after 5 epochs based on the validation accuracy on the development set. The search range and the optimal values for all hyperparameters are shown in Table 2.

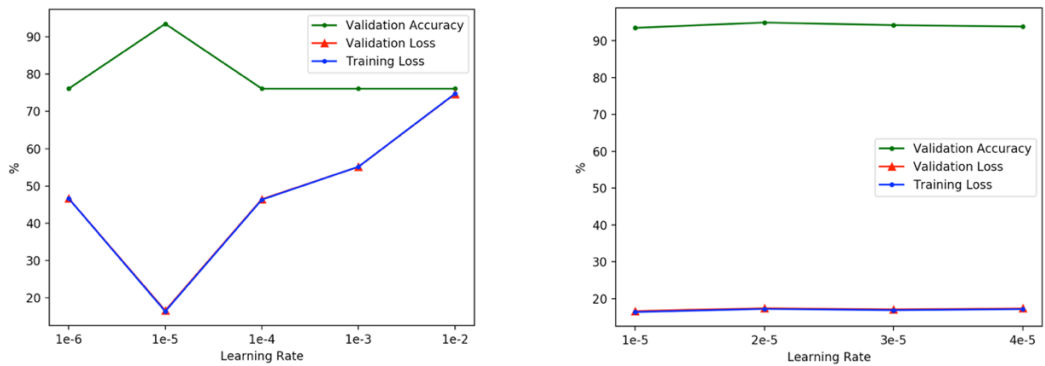

**Figure 3.** Grid search for learning rate using the development set with a larger step size (left) and then finer-grained step size (right)

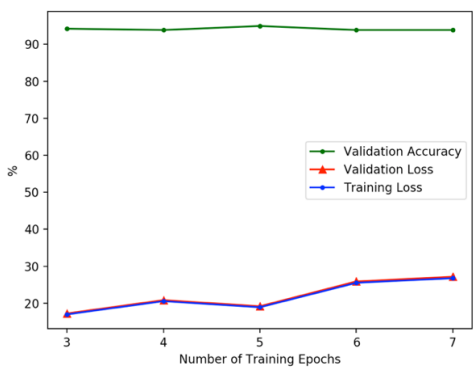

**Figure 4.** Finding the optimal number of train epochs based on the validation accuracy on the development set

**Table 2.** Hyperparameter systematic search range and their optimal values (marked in bold)

| Hyperparameter | Range |
|---|---|
| Max sequence length | (0,…, **128**,…, 400) |
| Batch size | 16, **32**, 64 |
| Learning rate | (1e-6,…, **2e-5**,…, 1e-2) |
| Number of training epochs | 3, 4, **5**, 6, 7 |

We pre-trained and fine-tuned the final BERT models on an NVIDIA Titan Xp graphics processing unit (GPU) using the optimal hyperparameters shown in Table 2. We pre-trained the BERT model for a total of 10,000 steps for 12 hours. Fine-tuning was computationally less expensive and took around 20 minutes on a single GPU and achieved a validation accuracy of 95% on the development set.

### Baseline Word2vec Model

*Word2vec Word Representations:* In order to evaluate the performance of the BERT contextual word representations in our task, we employed a pre-trained word2vec word representation model in our study as a baseline approach for comparison. The word2vec distributional semantics model is the most common method for generating word embeddings in NLP. Intuitively, the word2vec model is a neural network that maps words with similar context to nearby points in a vector space. However, once the word2vec training is completed, the same vector representation is used for all occurrences of a word in a corpus and the word context is not considered for modeling each occurrence. Our word2vec model was developed as part of our previous work to classify communication urgency in radiology reports [25]. This word2vec model utilized a skip-gram architecture with a vector representation dimension of 300 and a context window size of 8. We pre-trained the word2vec model on the entire text of the RadCore repository of 2 million radiology reports for 100,000 epochs or 24 hours on a single NVIDIA Titan Xp GPU.

*Training the Word2vec-Based Classifier:* We applied a fully-connected layer with a softmax normalization and a cross entropy loss function to build a classifier on top of the pre-trained word2vec word embeddings based on the labeled DHMC training dataset. To compute the final classification score we employed the following log softmax function: $P = \log(\text{softmax}(CW^T))$, where W is the classifier weights from the fully-connected layer, C is the aggregated vector representing a radiology report as we described in our previous work[25], and P is the confidence score of our binary classes indicating whether the radiology report is positive for prompt communication. We explored the optimal hyperparameters for the word2vec-based classifier on the development set, similar to the systematic search process that was described for the BERT model. Based on this systematic search, the optimal word2vec-based classifier had a batch size of 32, same as the BERT model, and the learning rate of 1e-4.

Of note, during the fine-tuning process, the BERT classifier was trained for a total of 10,000 epochs, and the word2vec-based classifier was trained for a total of 100,000 epochs. For both models, the optimal hyperparameters were determined based on the development dataset. Finally, the performance of the two optimized models were compared on the same evaluation set.

### Evaluation and Results

We used the held-out DHMC evaluation dataset to compare the performance of the BERT contextual word representations to the word2vec model as our baseline approach for predicting communication urgency based on radiology reports. In this evaluation, we measured standard classification evaluation metrics of precision, recall and F-measure to quantify the quality of the classification results. In addition, 95% confidence intervals were calculated for these metrics using the bootstrapping approach. These confidence intervals indicate the statistical power of these measures given the dataset size in this study.

For bootstrapping[28], we randomly selected *n* samples with replacement from the evaluation dataset, where *n* is the half of the size of the original evaluation dataset. We ran the bootstrapping for 1,000 iterations and calculated the average precision, recall and the F-measure and their 95% confidence intervals, as shown in Table 3.

**Table 3**. Performance metrics and their 95% confidence intervals (CIs) based on 1,000 iterations bootstrapping on the held-out evaluation dataset

| Model | Precision (%) | Recall (%) | F-measure (%) |
|---|---|---|---|
| Word2Vec-Based | 83.5 (76.6, 90.0) | 88.7 (81.6, 94.2) | 85.9 (80.6, 90.2) |
| BERT | **97.0** (93.3, 100.0) | **93.3** (87.9, 98.0) | **95.1** (91.7, 97.8) |

### Discussion

In this paper, we pre-trained and fine-tuned BERT, a contextual language model, on a large corpus of radiology reports to identify cases for prompt communication to the referring physicians. We employed the same training strategy as the BERT model's to build a classifier based on word2vec semantic representations. To maintain its consistency with our BERT model, we used the same training, development, evaluation dataset splits for training, hyperparameter tuning, and evaluation for the word2vect model. We compared the performance of the BERT model with the word2vec-based classifier, which is the previous state-of-the-art model for the identification of communication urgency based on radiology report texts.

Our evaluation showed that the BERT model achieved a precision of 97.0%, recall of 93.3%, and F-measure of 95.1% for identifying cases for prompt communication, outperforming the word2vec model, which had a precision of 83.5%, recall of 88.7%., and F-measure of 85.9%. Of note, the 95% confidence intervals of precision and F-measure for BERT and the word2vec model do not overlap. Therefore, the BERT's gain for identification of communication urgency in our evaluation set over the word2vec-based model based on these two metrics is statistically significant at the 0.05 level of significance.

We reviewed all the false positives and false negatives outputed by the BERT model in our evaluation set to conduct an error analysis for our proposed model. Through this error analysis, we observed commonalities in our errors that once remedied can help to improve the performance of the BERT model in future work. Most of the errors were caused due to the relatively long length of BERT input text from impression sections of the radiology reports in the evaluation set. As mentioned in the previous section, we treated the impression section of each radiology report as a single sequence of text as input to fine-tune the BERT model. Thus, we treated the whole impression section as a single sequence consisting of multiple subsentences. Of note, BERT in other classification applications, such as the CoLA task, was trained/ fine-tuned on single-sentences, rather than a long sequence of multiple sentences. The long length of the input sentences could have a negative impact on learning long-distance dependencies in the resulting contextual language model produced by the BERT model. In future work, we will investigate allternative attention mechasnims and skip-connections to adopt the BERT architecture for longer input sequences. In addition, we plan to make our input text more concise by eliminate less relvent sentence for our task. For example, we could leverage clinical text processing tools, such as NegEx[29], to eliminate normal and negative findings in the radiology report impression sections to place more emphasis on critical findings for our urgency classification task and make the input shorter and more appropriate for the BERT model.

It is widely known that lapses in the communication of medical findings, either due to delays or a lack of communication, increase the likelihood of adverse patient outcomes[30]. Particularly, high workload and lack of administrative support in radiology settings present challenges in identifying and communicating cases that require urgent management on the part of the referring physician[24]. Automatic methods that allow for accurate and rapid discrimination between cases requiring urgent communication and those that do not are highly beneficial. In this study, we developed a reliable and efficient model for identification of urgent cases that require prompt communication between a radiologist and a referring physician based on radiology reports. We expect the proposed approach in this paper can significantly contribute to the development of clinical decision support systems to automatically find and flag cases that require communication with referring physicians based on free-text radiology reports.

Our study has several limitations. Although the BERT model outperformed the word2vec-based model, it did not outperform the word2vec model with statistical significance on the recall metric. In addition, our evaluation dataset was from a single institution (i.e., DHMC). As future work, we plan to extend our evaluation to a larger dataset from multiple institutions to further investigate the performance of our contextual language model in analyzing radiology reports and to grasp its generalizability. Also, our proposed classification model in this work for the identification of urgency in radiology reporting only relies on the patients' radiology reports. In future work, we plan to incorporate additional related clinical information from electronic medical records to improve the accuracy of our model in assisting radiologists in detecting cases that require immediate attention by referring physicians.

As future work, we also plan to extend the BERT language model beyond radiology reports by pre-training BERT on other available large medical corpora. We also plan to apply the resulting contextual word embeddings in different biomedical applications. Particulary, we aim to utilize BERT pre-trained word embeddings in a multi-task learning framework to extract various clinically significant elements of information from biomedical text. Finally, our team will pursue implementation of the proposed approach as part of a clinical decision support system in PACS to facilitate the prompt communication of urgent findings between radiologists and referring physicians.

**Conclusion**

In this paper, we built a BERT contextual language model on a large corpus of radiology reports to identify radiology reports requiring prompt communication with the referring physicians. We compared the performance of BERT with the state-of-the-art word2vec model as the baseline approach. In our experiment, we used the same training and fine-tuning strategies to build two classifiers based on BERT and word2vec word embeddings for classification of communication urgency. Evaluation of the two models on our 425 radiology report evaluation dataset showed that the BERT model significantly outperformed the word2vec model on precision and F-measure. The clinical implementation of our proposed method could help improve patient outcomes by eliminating some cases of human error and expediting communication of urgent findings in radiology reports.

## Acknowledgments

The authors would like to thank Lamar Moss for his feedback on this paper. This work was supported in part by a grant from the US National Cancer Institute (R01CA249758).